\documentclass[letterpaper, 10 pt, conference]{ieeeconf}  

\IEEEoverridecommandlockouts                              

\usepackage{graphicx}
\usepackage{multirow}
\usepackage{array}
\usepackage{amssymb}
\usepackage{bbding}
\usepackage{booktabs}   
\usepackage{diagbox}
\usepackage{tabularx} 
\usepackage{ragged2e} 
\usepackage{amsmath}
\usepackage{xcolor} 
\usepackage{colortbl}
\usepackage{subcaption}
\usepackage{pifont}

\usepackage{caption}
\usepackage{makecell} 

\title{\LARGE \bf
RealMirror: A Comprehensive, Open-Source Vision-Language-Action Platform for Embodied AI
}

\author{
    Cong Tai \dag, \ 
    Zhaoyu Zheng \dag, \ 
    Haixu Long \dag, \ 
    Hansheng Wu, \ 
    Haodong Xiang, \ \\
    Zhengbin Long, \ 
    Jun Xiong, \ 
    Rong Shi, \ 
    Shizhuang Zhang, \ 
    Gang Qiu, \ 
    He Wang, \ \\
    Ruifeng Li, \  
    Jun Huang, \ 
    Bin Chang, \ 
    Shuai Feng, \ 
    Tao Shen* 
\thanks{\dag\ Equal contribution }
\thanks{* Corresponding author. Emails: shen.tao5@zte.com.cn}
\thanks{Cong Tai, Zhaoyu Zheng, Haixu Long, Hansheng Wu, Haodong Xiang, Zhengbin Long, Rong Shi, Shizhuang Zhang, Gang Qiu, He Wang, Ruifeng Li, Jun Huang, Bin Chang, Shuai Feng, Tao Shen are with ZTE Corporation, China.}
\thanks{Jun Xiong is with The Chinese University of Hong Kong, Shenzhen, China.}
}

\let\oldtwocolumn\twocolumn
\renewcommand\twocolumn[1][]{%
    \oldtwocolumn[{#1}{
    \begin{center}
        \vspace{-0.9cm} 
        \includegraphics[width=\textwidth]{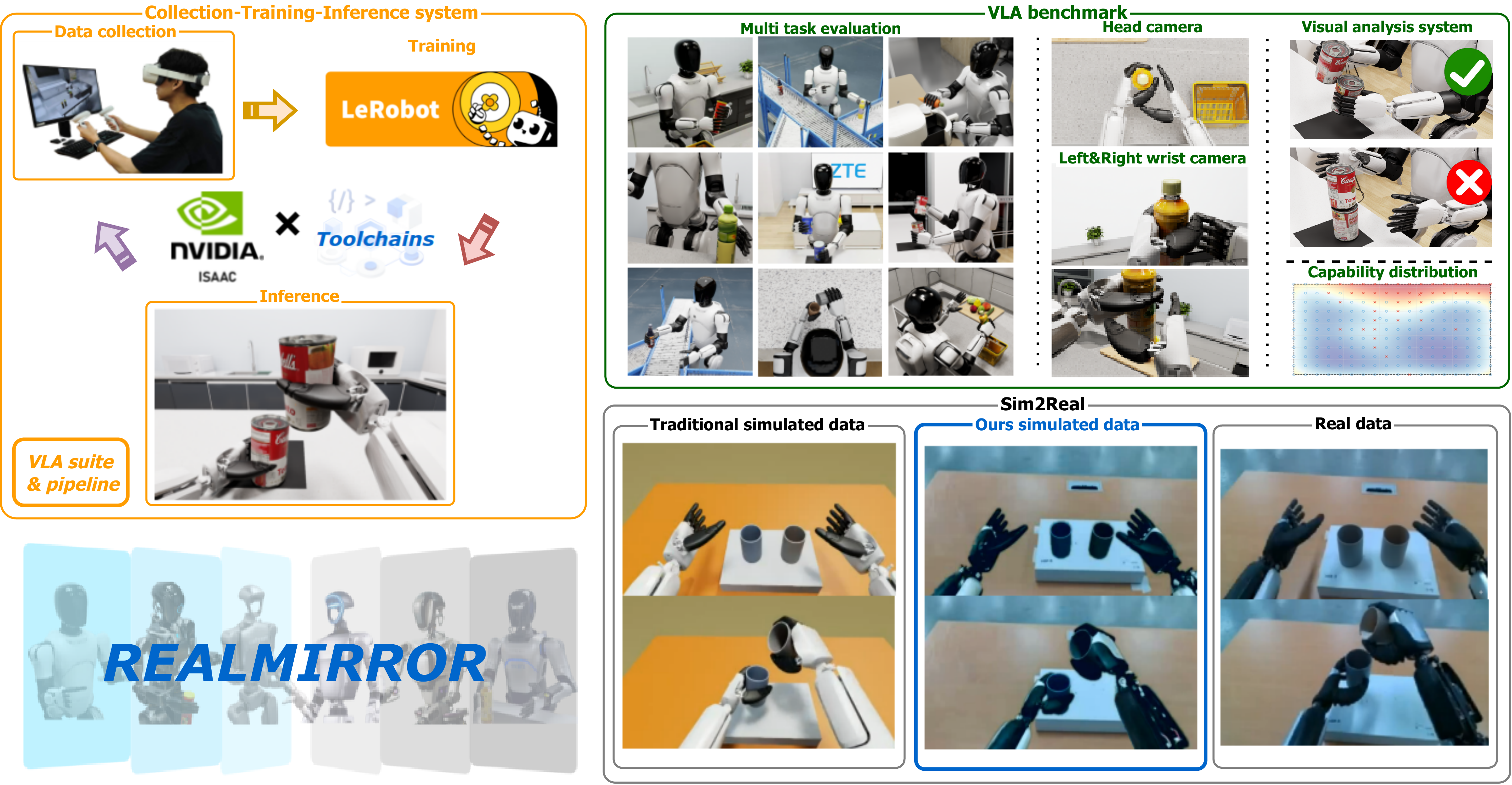}
        \captionof{figure}{The RealMirror framework for accelerating VLA research in humanoid robots. (a) Top-left: Our data collection, training, and inference system is developed based on VR teleoperation, Lerobot, and Isaac Sim. (b) Top-right: This platform populates a benchmark for humanoid robots with multiple scenarios and various VLA models. (c) Bottom-right: The striking photorealism of the simulation bridges the reality gap, enabling zero-shot Sim2Real transfer without any fine-tuning.}
        \label{fig:teaser}
    \end{center}
    }]
}

\begin{document}
\maketitle

\thispagestyle{empty}
\pagestyle{empty}

\begin{abstract}

The emerging field of Vision-Language-Action (VLA) for humanoid robots faces several fundamental challenges, including the high cost of data acquisition, the lack of a standardized benchmark, and the significant gap between simulation and the real world.
To overcome these obstacles, we propose RealMirror, a comprehensive, open-source embodied AI VLA platform. 
RealMirror builds an efficient, low-cost data collection, model training, and inference system that enables end-to-end VLA research without requiring a real robot. 
To facilitate model evolution and fair comparison, we also introduce a dedicated VLA benchmark for humanoid robots, featuring multiple scenarios, extensive trajectories, and various VLA models. 
Furthermore, by integrating generative models and 3D Gaussian Splatting to reconstruct realistic environments and robot models, we successfully demonstrate zero-shot Sim2Real transfer, where models trained exclusively on simulation data can perform tasks on a real robot seamlessly, without any fine-tuning. 
In conclusion, with the unification of these critical components, RealMirror provides a robust framework that significantly accelerates the development of VLA models for humanoid robots.
Project page: https://terminators2025.github.io/RealMirror.github.io

\end{abstract}
\section{INTRODUCTION}

The rapid evolution of Large Language Models (LLMs) like GPT \cite{achiam2023gpt}, Qwen \cite{yang2025qwen3}, and Deepseek \cite{liu2024deepseek} has significantly advanced the development of Artificial General Intelligence (AGI). While exhibiting remarkable model performance, they lack the ability to perform tasks in the real world. The vision of embodied AI can overcome this limitation by creating intelligent agents capable of perceiving, understanding, and physically interacting with the real world. The latest developments in humanoid robots and Vision-Language-Action (VLA) models are making this vision possible~\cite{zhao2023learning, chi2023diffusion, shukor2025smolvla}.

However, there are still a series of profound challenges that need to be addressed. First and foremost, the acquisition of high-quality interactive data remains an immense and costly bottleneck. Unlike large language models that can leverage vast internet datasets, embodied AI requires data generated from real robot interactions. This process is inherently time-consuming, expensive, and sometimes dangerous. Despite the availability of open-source robot datasets \cite{o2024open, bu2025agibot, wu2024robomind}, their offline nature prevents them from supporting the interactive iteration and validation necessary for embodied AI. Simulation platforms offer a promising alternative to address this data bottleneck, but existing platforms \cite{liu2023libero, mu2021maniskill, gu2023maniskill2, james2020rlbench} are primarily designed for robotic arms and grippers, lacking support for complex systems such as humanoid robots and dexterous hands. Furthermore, even platforms that support humanoid robots \cite{zhang2025agentworld} often suffer from insufficient environmental realism, which prevents a seamless Sim2Real transfer. This ``reality gap'' often results in suboptimal performance when models trained in simulation are deployed on real robots. Finally, the absence of a unified open-source humanoid robot benchmark for objectively evaluating and comparing model performance presents a significant obstacle in VLA, hindering systematic research.

To address these issues, we propose RealMirror, a comprehensive, open-source embodied AI VLA platform, as shown in Fig. \ref{fig:teaser}.
Firstly, to tackle the data acquisition and interactive validation bottleneck, we build an efficient, low-cost data collection, model training, and model inference system. We optimized the teleoperation and communication frameworks, which, compared to the general communication framework \cite{unitreerobotics_xr_teleoperate}, significantly enhanced the real-time performance and efficiency of data collection. When integrated into RealMirror, this enables end-to-end VLA research without the need for a real robot.
Secondly, we propose a dedicated VLA benchmark for humanoid robots to accelerate algorithm research and fair comparison. This benchmark includes five distinct scenarios and over 1,000 robot trajectories, designed to evaluate a suite of core competencies, from fundamental manipulations to dual-arm collaboration. Additionally, we conduct extensive automated evaluations on a variety of representative VLA models \cite{zhao2023learning, chi2023diffusion, shukor2025smolvla}.
Finally, to bridge the Sim2Real gap, we employ generative models \cite{hunyuan3d2025hunyuan3d} to create high-fidelity 3D assets from real-world images, and integrate 3D Gaussian Splatting (3DGS) \cite{moenne20243d} to reconstruct realistic environments and controllable robot models from video. Without any fine-tuning on real-world data, the model trained solely on simulation data can complete tasks seamlessly on a real robot.

In summary, our platform RealMirror enables researchers to perform data collection, model training, model inference, and performance evaluation in a unified system, thereby accelerating the development of VLA for humanoid robots. Our contributions are as follows:

1)	\textbf{We build an efficient, low-cost data collection, model training, and model inference system} that enables end-to-end VLA research without requiring a real robot.

2)	\textbf{We propose a dedicated VLA benchmark for humanoid robots} that facilitates model evolution and fair comparison through extensive experiments and automated evaluation across multiple scenarios and various VLA models.

3)	\textbf{We demonstrate the feasibility of zero-shot Sim2Real} by integrating generative models and 3DGS to reconstruct realistic environments and robot models, thus enabling models trained solely on simulation data to perform tasks seamlessly on a real robot without any fine-tuning.

\section{RELATED WORK}

\subsection{Vision-Language-Action}


Foundational works in this domain include RT-1 \cite{brohan2022rt}, a pioneering work that applied the Transformer architecture to robot control, and ACT \cite{zhao2023learning}, which employs a Variational Autoencoder for imitation learning. The development of RT-2 \cite{zitkovich2023rt} and OpenVLA \cite{kim2025openvla} further advanced this research by fine-tuning large-scale Vision-Language Models (VLMs), enabling robots to leverage web-scale knowledge. Notably, Diffusion Policy \cite{chi2023diffusion} introduced a policy learning method based on diffusion models, which is effective at handling high-dimensional and complex actions. 

Recent advancements have spurred the adoption of hierarchical \cite{shi2025hi, intelligence2504pi0} and dual-system architectures \cite{black2024pi_0, bjorck2025gr00t, shukor2025smolvla}. The Hi Robot framework \cite{shi2025hi}, for instance, leverages a pre-trained VLM as a high-level planner. Concurrently, a new wave of models—including PI0 \cite{black2024pi_0}, GROOT N1 \cite{bjorck2025gr00t} and SmolVLA \cite{shukor2025smolvla}—employs a dual-system architecture that combines a VLM for interpreting environments and instructions with a diffusion-based Transformer action expert for real-time action generation, representing the current frontier in embodied AI research.

\begin{figure*}[!ht]
    \centering
    \includegraphics[width=1.0\linewidth]{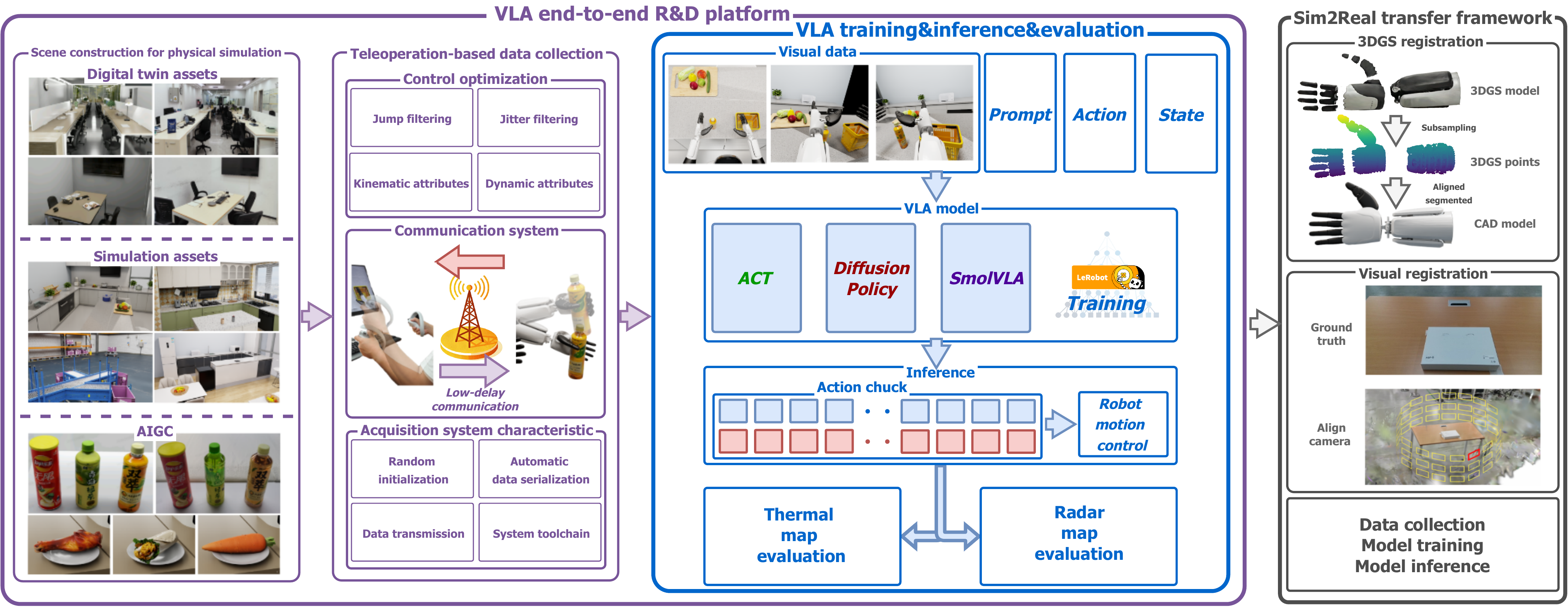}
    \caption{An overview of the integrated RealMirror pipeline. The platform offers an end-to-end solution for VLA research, encompassing asset acquisition and scene construction, as well as optimized teleoperation-based data collection and the training and evaluation of multiple VLA models. To bridge the critical reality gap, RealMirror incorporates a zero-shot Sim2Real module that uses 3D Gaussian Splatting to create a high-fidelity digital twin of the robot and the environment. This enables policies trained purely in simulation to be deployed directly to the real world.}
    \label{fig:scene_pipeline}
    \vspace{-0.6cm}
\end{figure*}

\subsection{Sim2Real}
Simulation has become an indispensable tool for training embodied agents like VLA models, as it circumvents the cost, inefficiency, and safety risks inherent to real-world data collection \cite{zhao2020sim, hofer2021sim2real}. This has spurred the development of numerous simulators to support large-scale robot learning \cite{todorov2012mujoco, mu2021maniskill, gu2023maniskill2}. However, the utility of simulation data is fundamentally limited by the Sim2Real gap \cite{zhao2020sim, abeyruwan2023sim2real}. Consequently, bridging this gap has been a central theme in robotics research, with significant progress in domains such as legged locomotion \cite{radosavovic2023learning, agarwal2023legged}, autonomous driving \cite{hu2023simulation}, and dexterous manipulation \cite{yuan2024robot, lum2024dextrah}.

The frontier of this research is zero-shot Sim2Real transfer, where policies trained entirely in simulation are deployed directly to hardware without fine-tuning. While recent works have demonstrated its feasibility for tasks like locomotion and mobile manipulation \cite{qureshi2024splatsim, liu2025fetchbot}, a critical gap persists for general-purpose, high-dimensional visuomotor policies. Humanoid robots equipped with dexterous hands represent a pinnacle of this challenge, demanding nuanced, whole-body control for contact-rich tasks. Our work directly addresses this challenge by introducing RealMirror, the first platform and benchmark designed to facilitate and evaluate zero-shot Sim2Real transfer for dexterous humanoid VLA policies, thereby catalyzing reproducible research in this ambitious domain of embodied AI.

\subsection{Simulation Platform}

Contemporary embodied robots are in urgent need of an interactive, high-fidelity simulation platform. Rcare world \cite{ye2022rcare}  is a high-fidelity, human-centric robotic caregiving simulation environment built with Unity. The ManiSkill series of works \cite{gu2023maniskill2,mu2021maniskill} focuses on manipulation skills over diverse objects in a full-physics simulator. In a similar vein, a benchmark for robotic learning is presented in Rlbench \cite{james2020rlbench}, which features a set of pre-defined tasks. Furthermore, Behavior-1k \cite{li2023behavior} and AgentWorld \cite{zhang2025agentworld} increase scene complexity by simulating human-like activities. Gaussian Splatting is also employed by Splatsim \cite{qureshi2024splatsim} to render real-world environments. This approach is a significant step towards bridging the Sim2Real gap. The availability of an open and user-friendly teleoperation platform for data collection is also of paramount importance. AgentWorld \cite{zhang2025agentworld} implements human teleoperation based on Isaac Sim \cite{isaacsim4.5}. Although these existing works have designed and realized high-fidelity, interactive simulation platforms, RealMirror distinguishes itself by demonstrating zero-shot generalization of dexterous hand skills to real-world scenarios, thereby validating the efficacy of data collection to real-world deployment.

\section{RealMirror}

\begin{table*}[!h]
\centering
\resizebox{\textwidth}{!}{  

\begin{tabular}{c|ccc|cc}
\toprule
\multirow{2}{*}{Name} & \multicolumn{3}{c}{\textbf{Data Collection}} & \multicolumn{2}{c}{\textbf{Platform Capabilities}} \\
\cmidrule(lr){2-4} \cmidrule(lr){5-6}
 & Tele-operation & Dexterous-hand & Num of Trajectories & End-to-end Framework & Zero-shot Sim2Real \\
\midrule

Maniskill2\cite{gu2023maniskill2}       & \XSolidBrush & \XSolidBrush & 30k    & \XSolidBrush & \XSolidBrush \\
RLBench\cite{rlbench}             & \XSolidBrush & \XSolidBrush & --     & \XSolidBrush & \XSolidBrush \\
BiGym\cite{bigym}                 & \Checkmark   & \XSolidBrush & $>$2000 & \XSolidBrush & \XSolidBrush \\
Behavior-1K\cite{behavior1k}      & \XSolidBrush & \XSolidBrush & --     & \XSolidBrush & \XSolidBrush \\
MimicGen\cite{mimicgen}           & \XSolidBrush & \XSolidBrush & 50k    & \XSolidBrush & \XSolidBrush \\
GRUtopia\cite{grutopia}           & \Checkmark   & \XSolidBrush & --     & \XSolidBrush & \XSolidBrush \\
AgentWorld\cite{zhang2025agentworld} & \Checkmark & \Checkmark   & $>$1000 & \XSolidBrush & \XSolidBrush \\
\midrule
RealMirror (Ours)                 & \Checkmark   & \Checkmark   & $>$1000 & \Checkmark   & \Checkmark \\
\bottomrule
\end{tabular}
}
\vspace{0.1cm}
\caption{Comparison of robotic simulation platforms in terms of data collection methods and platform-level capabilities. 
RealMirror distinguishes itself by providing a complete, end-to-end framework that supports the full VLA research lifecycle, from data collection to direct real-world deployment.}
\label{tab:sim_compare}
\vspace{-0.5cm}
\end{table*}

RealMirror is a comprehensive, open-source embodied AI platform designed for humanoid VLA research. 
It integrates an efficient, low-cost system for data collection, model training, and inference, enabling end-to-end VLA development without requiring a real robot. 
To facilitate model evolution, RealMirror provides a dedicated benchmark consisting of over 1,000 high-quality simulation trajectories across multiple tasks and humanoid robot platforms. 
Finally, the platform incorporates a Sim2Real transfer framework, leveraging generative models \cite{hunyuan3d2025hunyuan3d} and 3DGS \cite{wu20253dgut, kerbl20233d} to reconstruct realistic environments and robot models, thereby enabling zero-shot transfer from simulation to real-world execution. 

Table~\ref{tab:sim_compare} summarizes the key features of RealMirror compared with existing robotic simulation platforms, and Fig.~\ref{fig:scene_pipeline} provides an overview of the platform architecture.

\subsection{Scene Construction for Physical Simulation} 
To support diverse embodied AI tasks, we construct simulation environments based on the NVIDIA Isaac Sim platform~\cite{isaacsim4.5}. 
Specifically, we design a wide range of indoor scenes that incorporate complex layouts, multiple manipulable objects, and realistic physical interactions. 
By integrating CAD models and assets from various asset libraries~\cite{isaacsim4.5, jin2025artviparticulateddigitalassets, ycb}, RealMirror enables the creation of customizable environments with varying difficulty levels and task requirements. 
In addition, we assign appropriate physical properties (e.g., mass, friction, and collision parameters) to the assets, ensuring their plausibility in simulation and compatibility with humanoid robot embodiments. 
These simulation-ready environments serve as the foundation for large-scale data collection, training, and evaluation in RealMirror. 

\subsection{Efficient System for Data Collection, Training, and Inference}
We have developed an efficient, end-to-end system for VLA data collection, training, and inference. 
By leveraging a teleoperation system, the platform enables high-quality data acquisition, while the unified training and inference framework supports multiple state-of-the-art VLA models. 
Deep integration with Isaac Sim allows for closed-loop evaluation, ensuring both efficiency and reproducibility. 
Overall, this system significantly reduces the cost and complexity of data collection and provides a reliable foundation for subsequent VLA research and model benchmarking.
\subsubsection{Teleoperation-Based Data Collection}


We develop a teleoperation-based data collection system to efficiently gather high-quality training data for humanoid VLA tasks. 
The system consists of two main components. 
First, we implement a motion control pipeline in the simulation environment with multi-level filtering mechanisms, including: IK joint control jump filtering, end-effector pose communication and drift compensation, IK solver threshold filtering, and cross-frame end-effector pose threshold filtering. 
These layers of filtering ensure smooth and physically plausible motion during teleoperation. 
Second, we build a lightweight WebXR-based communication system for real-time control. By streamlining the data transmission protocol, our system operates at a 90 Hz transmission frequency and achieves a 114ms reduction in end-to-end latency from teleoperation command to robot execution, compared to the general communication framework \cite{unitreerobotics_xr_teleoperate}.

Together, these components enable efficient data acquisition. 
For example, a single-arm pick-and-place task averages 7.83 seconds per trajectory, encompassing the full end-to-end workflow: environment initialization, object manipulation, and data packaging for subsequent training.

\subsubsection{Unified Training and Inference Framework} 


Our framework provides a unified training and inference system for VLA research. 
On the training side, we support multiple representative VLA models, including ACT~\cite{zhao2023learning}, Diffusion Policy~\cite{chi2023diffusion}, and SmolVLA~\cite{shukor2025smolvla}. In addition, all our algorithms are adapted to incorporate a temporal ensembling mechanism to enhance action prediction robustness and reduce compounding errors.
To ensure a fair and efficient comparison, we build upon the LeRobot library \cite{cadene2024lerobot} and extend its functionalities to support humanoid robot embodiments. 
On the inference side, the trained models are integrated with Isaac Sim \cite{isaacsim4.5} for interactive evaluation. 
During inference, the system continuously receives multi-modal inputs, including BGR observations, robot proprioceptive states, and natural language instructions. 
The VLA model then predicts the corresponding action sequences, which are published back to the simulator for real-time execution. 
This closed-loop integration enables seamless evaluation of different models under consistent conditions, thereby facilitating fair benchmarking and systematic analysis.


\subsection{Benchmark for Humanoid VLA} 



We introduce a systematic benchmark for humanoid robot VLA research.
The benchmark provides a standardized platform for training, evaluating, and comparing VLA algorithms across a wide range of tasks and scenarios. 
It comprises over 1,000 high-quality simulated trajectories, capturing diverse interactions, manipulation challenges, and dynamic environments. 
By combining task diversity, multi-modal inputs, and a unified evaluation pipeline, this benchmark enables systematic, reproducible, and rigorous research on humanoid VLA.

\subsubsection{Task Scenarios and Dataset}

We construct a high-quality humanoid VLA dataset comprising five task scenarios, each with 240 trajectories, totaling over 1,000 simulated trajectories. 
The task scenarios are designed to cover a broad range of skills, including  \textit{Pick and Place}, \textit{Dual-arm collaboration}, \textit{Push and Pull}, \textit{Dynamic grasping}, and \textit{Precision control}.
Table~\ref{tab:benchmark_dataset} presents the distribution of demonstration trajectories in the benchmark dataset for the humanoid robot. 
Below, we describe each task scenario:

\begin{itemize}
    \item \textit{\textbf{Kitchen Cleanup}}: Use the left hand to pick up chips, green tea, or lemon tea from the table, then transfer the item to the right hand and place it into the basket.
    \item \textit{\textbf{Air Fryer Manipulation}}: Use the left hand to lift chicken rolls, chicken legs, or carrots from a plate, then use the right hand to open the air fryer and place the food inside.
    \item \textit{\textbf{Assembly Line Sorting}}: Sort three types of items (oil, cola, and Sprite) on the conveyor belt, ensuring that each sorted item lands correctly in its designated box.
    \item \textit{\textbf{Cup-to-Cup Transfer}}: Pour berries from the cup on the right into the cup on the left.
    \item \textit{\textbf{Can Stacking}}: Stack cans from both sides into the center and ensure they are placed stably.
\end{itemize}

\begin{table}[h]
    \centering
    \caption{Mapping of benchmark tasks to the core skills they assess, and the corresponding number of demonstration trajectories.}
    
    \label{tab:benchmark_dataset}
    \begin{tabularx}{\columnwidth}{l >{\raggedright\arraybackslash}X c}
        \toprule
        \textbf{Task} & \textbf{Assessed Skills} & \textbf{Num of Traj.} \\
        \midrule
        Kitchen Cleanup & 
        \makecell[l]{- Pick and Place \\ - Dual-arm collaboration} & 240 \\
        \addlinespace 
        Air Fryer Manipulation & 
        \makecell[l]{- Pick and Place \\ - Push and Pull \\ - Dual-arm collaboration} & 240 \\
        \addlinespace
        Assembly Line Sorting & 
        \makecell[l]{- Pick and Place \\ - Dual-arm collaboration \\ - Dynamic grasping} & 240 \\
        \addlinespace
        Cup-to-Cup Transfer & 
        \makecell[l]{- Dual-arm collaboration \\ - Precision control} & 240 \\
        \addlinespace
        Can Stacking & 
        \makecell[l]{- Pick and Place \\ - Precision control} & 240 \\
        \midrule
        \textbf{Total} & & \textbf{1200} \\
        \bottomrule
    \end{tabularx}
    \vspace{-0.5cm}
\end{table}

\subsubsection{Supported VLA Models}

We train several representative VLA algorithms on our benchmark, covering different design philosophies:
\textbf{ACT}~\cite{zhao2023learning} is based on a VAE architecture for VLA modeling, emphasizing fine-grained bimanual manipulation.  
\textbf{Diffusion Policy}~\cite{chi2023diffusion} predicts actions via a generative diffusion process, offering robust visuomotor control.  
\textbf{SmolVLA}~\cite{shukor2025smolvla} employs a dual-system architecture, where S2 VLM extracts visual and textual information and S1 Action Expert predicts corresponding actions, enabling broad generalization.  

\subsubsection{Evaluation Metrics}
Evaluation metrics focus on task success rate, which provides a clear and consistent measure of VLA performance across different tasks and scenarios. 
We provide automated evaluation tools to ensure objective, reproducible, and fair assessment of all models, enabling standardized benchmarking for future research.
Simultaneously, to intuitively analyze how the capabilities of different models are distributed across the same scenario, we developed a heatmap analysis tool, as depicted in Fig. \ref{fig:teaser}.

\subsection{Sim2Real Transfer Framework}

While Isaac Sim provides a physically robust foundation, bridging the visual gap between simulation and reality is crucial for effective policy transfer. 
We employ a multi-pronged strategy that ensures the VLA model observes a simulation environment virtually indistinguishable from the real world. 
By integrating high-fidelity background reconstruction, photorealistic robot modeling, and differentiated interactive object generation, our pipeline enables zero-shot transfer of trained VLA policies to real-world deployment.


\subsubsection{Visual Augmentation for Realism}
We adopt a hybrid generative approach to enhance visual realism. 
Different scene components, including the static background, the robot, and interactive objects, are treated with specialized techniques to ensure high-fidelity perception.


\textbf{Static Environment Rendering with 3DGS:} 
We capture the target real-world workspace from multiple viewpoints and reconstruct the entire static scene using 3DGS. 
This high-fidelity background, including lab benches, walls, and distant clutter, is integrated into the simulator as a non-interactive canvas, preserving subtle lighting and material effects that are difficult to reproduce with standard rendering.


\textbf{High-Fidelity Articulated Robot Model:} 
We reconstruct the physical humanoid robot with 3DGS and segment it into individual links. 
Each link is rigidly aligned to its corresponding USD model in Isaac Sim using a scale $S$, rotation $R$, and translation $T$ transformation:

\vspace{-0.2cm}
\begin{equation}
\vspace{-0.15cm}
P_{\text{USD}} = S \cdot R \cdot P_{\text{3DGS}} + T
\end{equation}

This process overlays a photorealistic "skin" onto the kinematically and physically accurate robot skeleton, ensuring visual fidelity in simulation.


\subsubsection{Differentiated Strategy for Interactive Objects}
Interactive objects are treated according to their physical and visual requirements:

\textbf{High-Precision Objects:} Objects requiring precise contact physics, such as dexterous end-effectors or flat tabletops, use a digital twin approach. A clean CAD model governs all dynamics and collisions, while a 3DGS reconstruction provides visual realism aligned to the CAD model.

\textbf{Low-Precision Objects:} Objects with lower physics requirements prioritize visual diversity. Using few-shot 3D generative models, we produce textured 3D meshes from a few images, which serve as both visual and collision representations, enabling rapid scene population with diverse assets.


\subsubsection{Coordinate Alignment and Camera Calibration}
We align the coordinate systems of Isaac Sim, the 3DGS reconstructed scene, and the robot's real-world pose to ensure consistency between the physical and simulated environments.
First, we employ the Iterative Closest Point (ICP) algorithm to align the CAD assets in Isaac Sim with the 3DGS reconstructed environment.
For camera calibration, we record the robot’s observed images $I_\text{robot}$ during task execution and solve for the camera pose using Structure-from-Motion (SfM) together with a set of real-world images.
Because the Isaac Sim coordinates are already aligned with the 3DGS scene, placing a virtual camera at the solved $I_\text{robot}$ pose within Isaac Sim achieves accurate calibration between the real and simulated cameras.

In summary, our Sim2Real transfer framework combines high-fidelity visual augmentation, precise robot modeling, and differentiated interactive object treatment to bridge the gap between simulation and reality. 
With coordinate alignment and camera calibration, trained VLA policies can be directly deployed on real robots without additional fine-tuning. 
This approach provides a robust foundation for efficient and scalable humanoid VLA research, ensuring that learned policies are both transferable and reliable across diverse scenarios.





\section{experiments}

The experiment consists of two main phases. First, we established a VLA benchmark by training and automatically evaluating various VLA models on our platform to compare their performance. Second, our Sim2Real experiments assess the effectiveness of models trained on simulation data when deployed on a real robot.

\subsection{VLA Benchmark}

\subsubsection{Experimental Setup}
Our datasets were collected using a PICO Neo3 Pro headset and a workstation with Ada5880.
To comprehensively evaluate our benchmark, we selected three representative VLA models for the experiments: ACT \cite{zhao2023learning}, Diffusion Policy \cite{chi2023diffusion}, and SmolVLA \cite{shukor2025smolvla}. All models process synchronized multi-view BGR images, while SmolVLA additionally utilizes natural language task descriptions. The unified action space is 26-dimensional, with 13 dimensions for each of the two robotic arms (7 for the arm and 6 for the hand). 
For training, each model was trained for 100,000 steps with a batch size of 16. Additionally, a Temporal Ensembler Mechanism was adopted to enhance the smoothness of actions.


\subsubsection{Automatic Evaluation Protocol}







The performance of models was quantitatively evaluated by their success rate across five distinct manipulation scenarios. The criteria for each scenario were defined as follows.

\begin{itemize}
\item \noindent \textit{\textbf{Kitchen Cleanup}} (400 trials): 
The task is successful when a specified item is picked up and placed into the designated basket by coordinating both of its arms.
\item \noindent \textit{\textbf{Air Fryer Manipulation}} (400 trials): 
A successful trial involves the robot correctly opening the air fryer drawer, placing a food item inside, and then closing the drawer. 
\item \noindent \textit{\textbf{Can Stacking}} (400 trials): 
This scenario requires the robot to grasp two cans on the desk and stably stack them employing a bimanual manipulation approach.
\item \noindent \textit{\textbf{Cup-to-Cup Transfer}} (200 trials): 
A trial is deemed successful upon the transfer of a berry from the right cup to the left cup, performed while the cups are lifted in the air. 
\item \noindent \textit{\textbf{Assembly Line Sorting}}  (100 trials): 
The criterion for successful completion is the correct sorting of three consecutive items from a conveyor belt in a single trial.
\end{itemize}

\subsubsection{Experimental Results}

\begin{table}[t]
    \centering
    \caption{Comparison of task success rates (\%) for each model. The best performance in each task is highlighted in \textbf{bold}.}
    \label{tab:vla_performance}
      \setlength{\tabcolsep}{3pt}
    \begin{tabular}{@{}l c c c@{}}
        \toprule
        {\textbf{Task}} & {\textbf{ACT} \cite{zhao2023learning}} & {\textbf{Diffusion Policy} \cite{chi2023diffusion}} & {\textbf{SmolVLA} \cite{shukor2025smolvla}} \\
        \midrule
        Kitchen Cleanup        & \textbf{100.00}  & 99.00             & 99.75 \\
        Air Fryer Manipulation & 77.75         & \textbf{85.50}  & 83.00             \\
        Assembly Line Sorting  & \textbf{95.00}   & 88.00             & 86.00             \\
        Cup-to-Cup Transfer    & 55.50          & 63.50           & \textbf{68.00}    \\
        Can Stacking           & 39.50         & 39.75          & \textbf{62.00}    \\
        \midrule 
        \textbf{Avg}           & 73.55          & 75.15          & \textbf{79.75} \\
        \bottomrule
    \end{tabular}
\vspace{-0.5cm}
\end{table}
The evaluation results for the three models are summarized in Table \ref{tab:vla_performance}, showing their success rates across the five scenarios.

From a task-centric perspective, the results show that while SmolVLA achieved the highest average success rate (79.75\%), its primary advantage was in high-precision scenarios like \textit{Cup-to-Cup Transfer} (68.00\%) and \textit{Can Stacking} (62.00\%). In contrast, ACT demonstrated near-perfect performance in \textit{Kitchen Cleanup} (100.00\%) and the dynamic \textit{Assembly Line Sorting} (95.00\%), while Diffusion Policy's strength was most apparent in \textit{Air Fryer Manipulation} (85.50\%). This comparative analysis reveals that while all three models possess domain-specific strengths, SmolVLA achieved the most robust performance across the evaluated tasks.

\begin{figure}[!ht]
    \centering
    \includegraphics[width=0.7\linewidth]{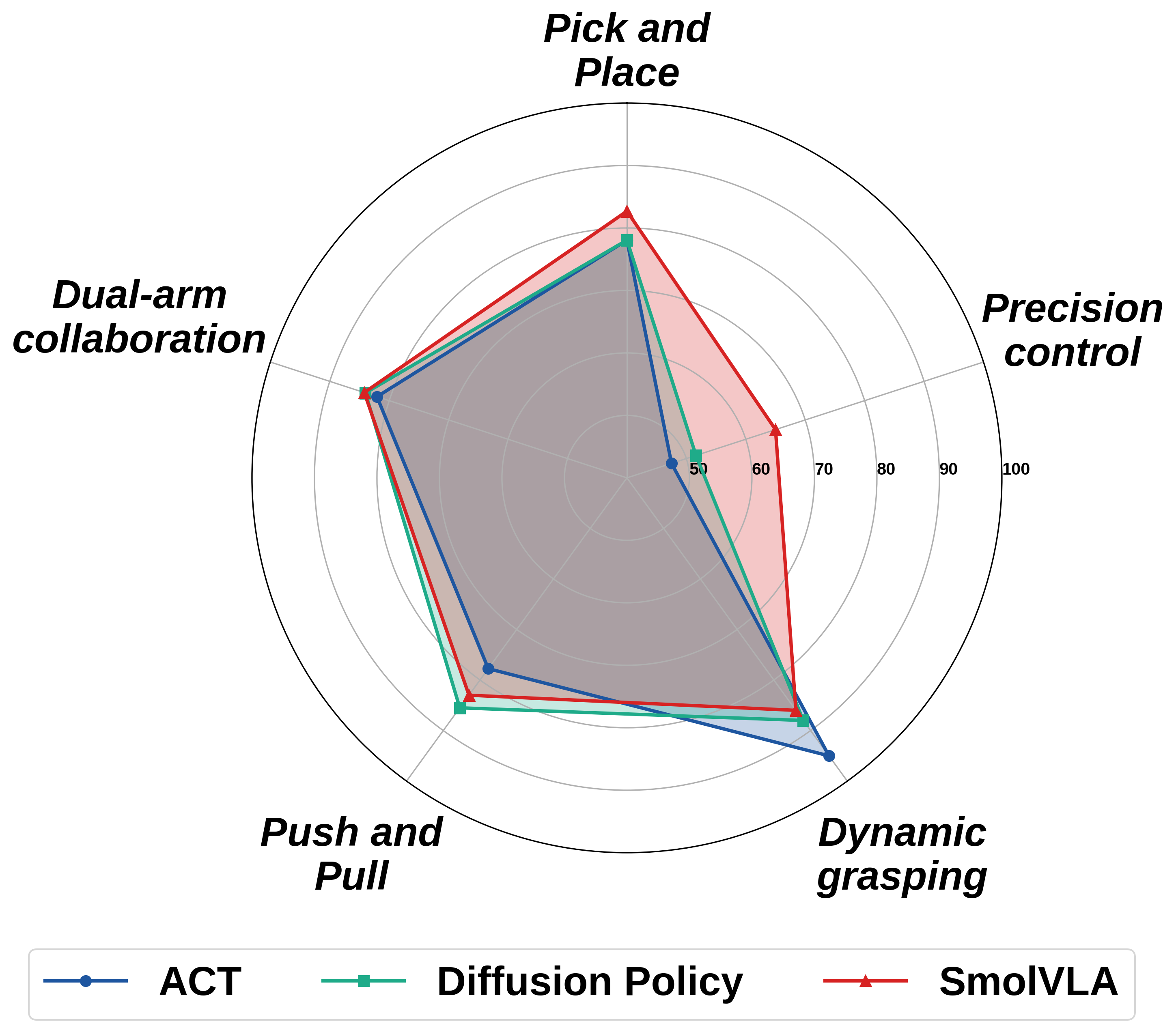}
    \caption{Model performance comparison across different robotic skills.}
    \label{fig:model_performance_radar}
    \vspace{-0.5cm}
\end{figure}

\begin{figure}[!ht]
    \centering
    \includegraphics[width=1.0\linewidth]{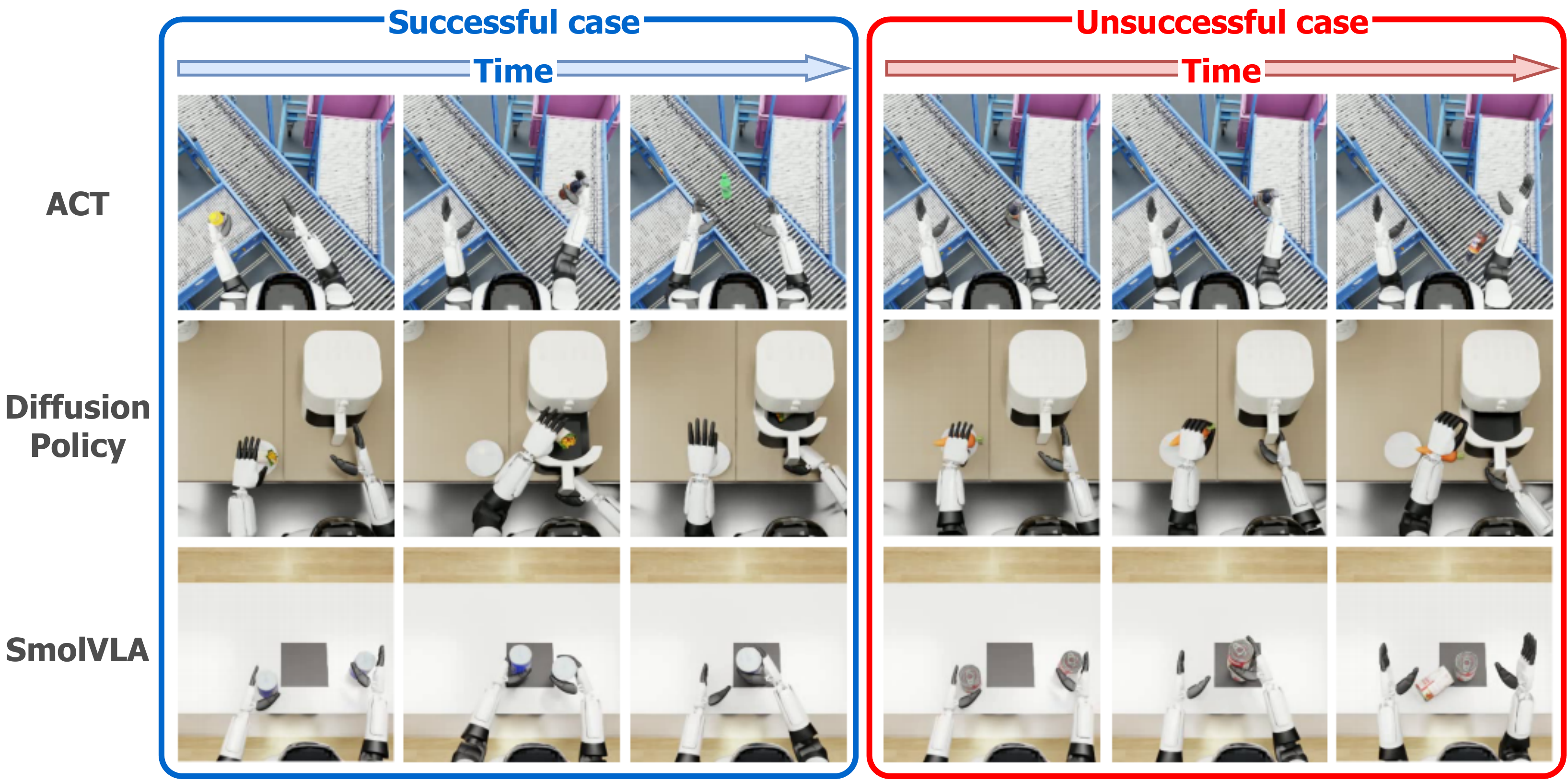}
    \caption{Qualitative results for different VLA algorithms in our benchmark.}
    \label{fig:benchmark_case}
    \vspace{-0.6cm}
\end{figure}

In addition to evaluating the models based on task completion success rates, our benchmark also analyzed their performance from a skill-based perspective. For this, we abstracted five core robotic skills: Pick and Place, Dual-arm collaboration, Push and Pull, Dynamic grasping, and Precision control. This approach provides a finer-grained understanding of each model, as shown in Fig. \ref{fig:model_performance_radar}. The analysis reveals that SmolVLA exhibited the most balanced performance, achieving the highest average success rates in both Pick and Place and Precision Control. The ACT model showed a significant advantage in Dynamic Grasping, with a 95\% success rate that was notably higher than the other two models. Meanwhile, Diffusion Policy performed best in the Push and Pull skill. Several qualitative results for the different VLA algorithms are shown in Fig.~\ref{fig:benchmark_case}.

This two-tiered evaluation framework, assessing performance at both the task and skill levels, is the cornerstone of the RealMirror benchmark's utility. It allows us to directly correlate a model's success in a complex task with its proficiency in underlying robotic skills. By providing this deeper, diagnostic insight, our benchmark empowers the community to move beyond simple leaderboards. It offers a systematic foundation for comparing VLA models, identifying their specific architectural trade-offs, and ultimately guiding future research toward targeted algorithmic improvements.

\subsection{Sim2Real Experiments}

\begin{figure}[!ht]
    \centering
    \includegraphics[width=1.0\linewidth]{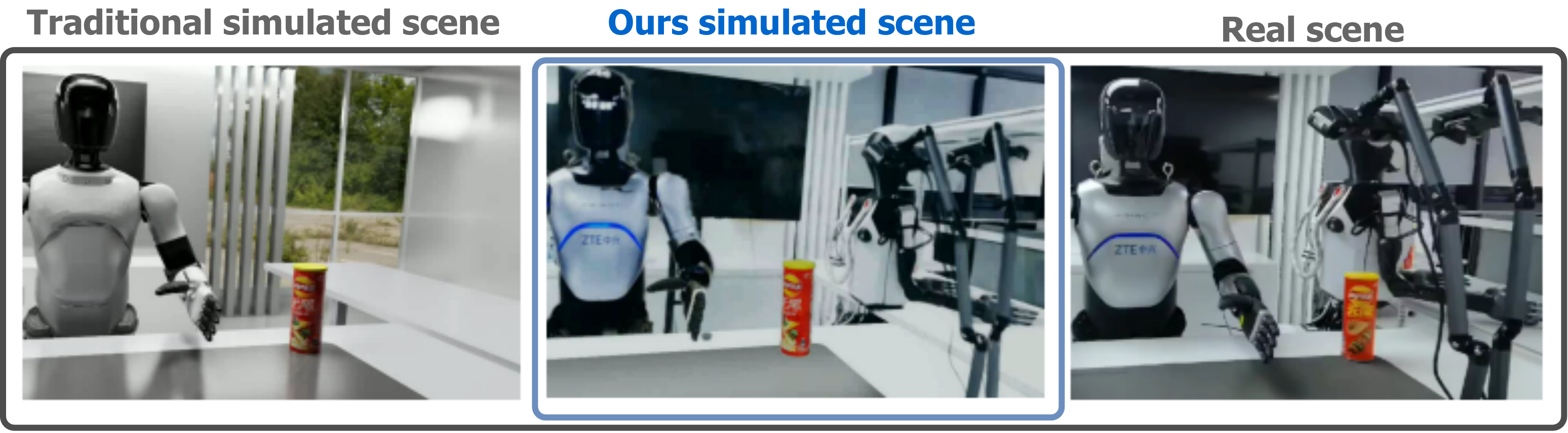}
    \caption{Comparison of traditional simulated, real, and our simulated scene.}
    \label{fig:sim_data_compare}
    \vspace{-0.6cm}
\end{figure}

Traditional robotic simulations suffer from a substantial Sim2Real gap, primarily in visual fidelity. This gap stems from discrepancies in lighting, material properties, textures, and geometry between the simulated and real worlds, which severely restricts the direct transferability of models trained on simulation data to real robots. To address this challenge, we integrated generative models and 3DGS to reconstruct highly realistic environments and robot models. The real-world experiments were implemented with a ZHIYUAN A2 robot. As illustrated in Fig. \ref{fig:sim_data_compare}, our reconstructed scenes are visually more realistic than those generated by traditional simulations and closely approximate real-world visual fidelity. This visual realism provides a robust foundation for zero-shot Sim2Real transfer.
We chose the ACT algorithm as our representative model because of its demonstrated efficiency and strong real-time inference performance. 
Meanwhile, we selected two distinct tasks: 
1) picking a chip from the edge of a table and placing it in the center. 
2) Pouring a ball from a cup held in the right hand into a cup held in the left. This is a simplified version of the Cup-to-Cup Transfer task, where we replaced the berry with a ball to reduce the randomness of the movement of the object.
For each task, a dataset was collected and utilized to train the models within the simulation environment. These trained models were subsequently evaluated on both the simulation platform and a real robot. 

\begin{figure}[!ht]
    \centering
    \includegraphics[width=1.0\linewidth]{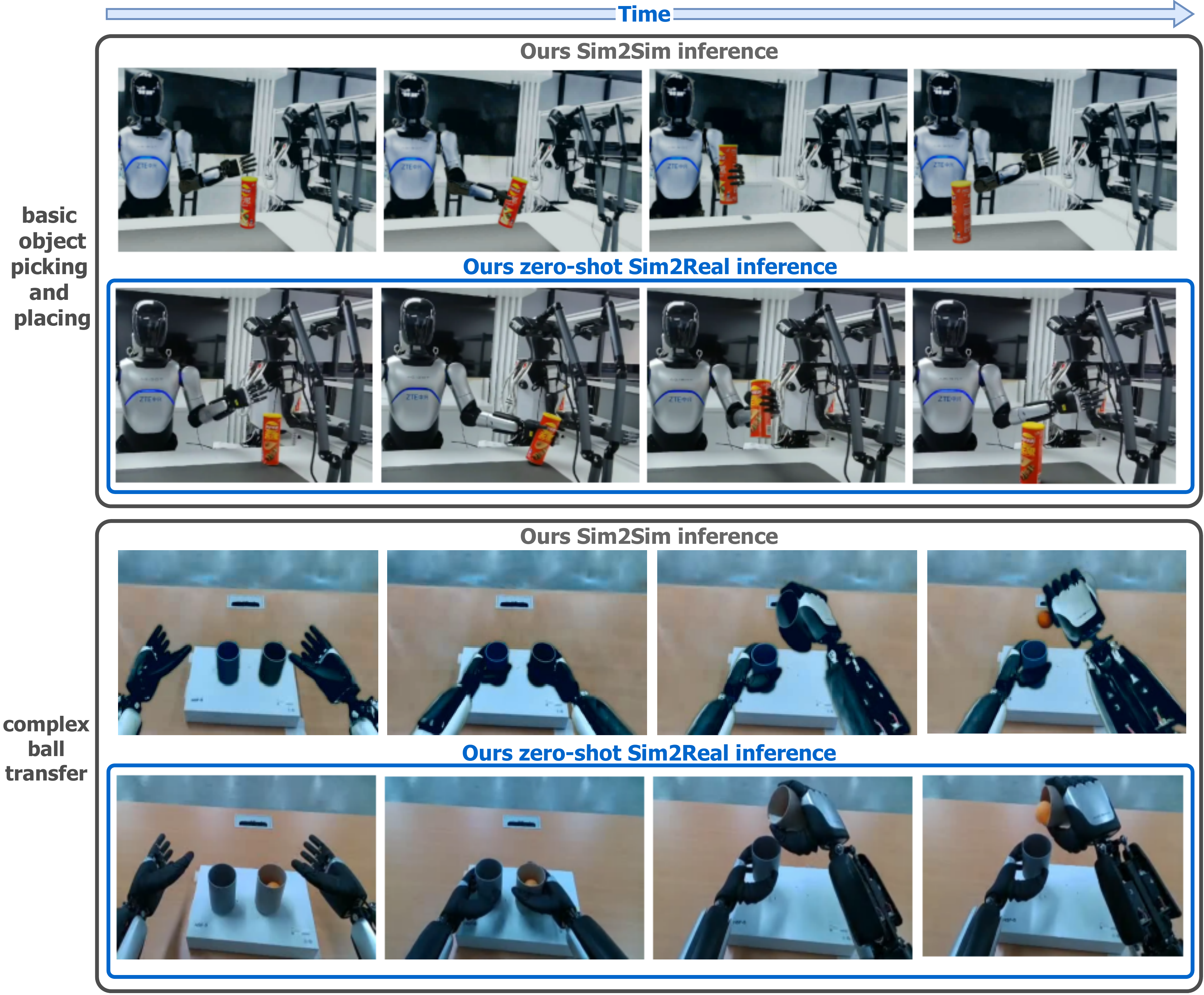}
    \caption{Comparison of Sim2Sim and Sim2Real inference results.}
    \label{fig:zero_shot}
    \vspace{-0.6cm}
\end{figure}

The results are shown in Fig. \ref{fig:zero_shot}. The models successfully executed both tasks with smooth and stable movements, from basic object picking and placing to more complex operations like dual-arm collaborative ball transfer. In addition to qualitative results, we also conducted quantitative experiments. 
The model achieved a 92.86\% accuracy on the basic object picking and placing task, while it reached 71.43\% Sim2Real accuracy for the complex ball transfer task without fine-tuning.


In summary, models trained exclusively on our simulation data can perform tasks seamlessly on a real robot without any fine-tuning, demonstrating the feasibility of zero-shot Sim2Real transfer.

\section{conclusions}

In response to the urgent necessity for a professional R\&D platform in humanoid robotics and dexterous manipulation, we introduce RealMirror, a comprehensive, open-source platform. This platform provides a seamless, end-to-end solution for data collection, model training, model inference, and performance evaluation. Meanwhile, we demonstrate that models trained in our high-fidelity simulation environment can achieve zero-shot Sim2Real transfer to the real world. We are currently advancing the platform by developing a novel automatic data collection method and expanding its simulation scenarios, which will enable the generation of large-scale, high-quality simulation data. Additionally, further zero-shot Sim2Real transfer experiments are being conducted to validate the universality and robustness of our platform across a wider range of scenarios. We believe these systematic advancements will significantly accelerate the development of embodied AI.


\bibliographystyle{IEEEtran}  
\bibliography{IEEEfull}     

\begin{thebibliography}{10}
\providecommand{\url}[1]{#1}
\csname url@rmstyle\endcsname
\providecommand{\newblock}{\relax}
\providecommand{\bibinfo}[2]{#2}
\providecommand\BIBentrySTDinterwordspacing{\spaceskip=0pt\relax}
\providecommand\BIBentryALTinterwordstretchfactor{4}
\providecommand\BIBentryALTinterwordspacing{\spaceskip=\fontdimen2\font plus
\BIBentryALTinterwordstretchfactor\fontdimen3\font minus \fontdimen4\font\relax}
\providecommand\BIBforeignlanguage[2]{{%
\expandafter\ifx\csname l@#1\endcsname\relax
\typeout{** WARNING: IEEEtran.bst: No hyphenation pattern has been}%
\typeout{** loaded for the language `#1'. Using the pattern for}%
\typeout{** the default language instead.}%
\else
\language=\csname l@#1\endcsname
\fi
#2}}

\bibitem{achiam2023gpt}
J.~Achiam, S.~Adler, S.~Agarwal, L.~Ahmad, I.~Akkaya, F.~L. Aleman, D.~Almeida, J.~Altenschmidt, S.~Altman, S.~Anadkat, \emph{et~al.}, ``Gpt-4 technical report,'' \emph{arXiv preprint arXiv:2303.08774}, 2023.

\bibitem{yang2025qwen3}
A.~Yang, A.~Li, B.~Yang, B.~Zhang, B.~Hui, B.~Zheng, B.~Yu, C.~Gao, C.~Huang, C.~Lv, \emph{et~al.}, ``Qwen3 technical report,'' \emph{arXiv preprint arXiv:2505.09388}, 2025.

\bibitem{liu2024deepseek}
A.~Liu, B.~Feng, B.~Xue, B.~Wang, B.~Wu, C.~Lu, C.~Zhao, C.~Deng, C.~Zhang, C.~Ruan, \emph{et~al.}, ``Deepseek-v3 technical report,'' \emph{arXiv preprint arXiv:2412.19437}, 2024.

\bibitem{zhao2023learning}
T.~Z. Zhao, V.~Kumar, S.~Levine, and C.~Finn, ``Learning fine-grained bimanual manipulation with low-cost hardware,'' \emph{arXiv preprint arXiv:2304.13705}, 2023.

\bibitem{chi2023diffusion}
C.~Chi, Z.~Xu, S.~Feng, E.~Cousineau, Y.~Du, B.~Burchfiel, R.~Tedrake, and S.~Song, ``Diffusion policy: Visuomotor policy learning via action diffusion,'' \emph{The International Journal of Robotics Research}, p. 02783649241273668, 2023.

\bibitem{shukor2025smolvla}
M.~Shukor, D.~Aubakirova, F.~Capuano, P.~Kooijmans, S.~Palma, A.~Zouitine, M.~Aractingi, C.~Pascal, M.~Russi, A.~Marafioti, \emph{et~al.}, ``Smolvla: A vision-language-action model for affordable and efficient robotics,'' \emph{arXiv preprint arXiv:2506.01844}, 2025.

\bibitem{o2024open}
A.~O’Neill, A.~Rehman, A.~Maddukuri, A.~Gupta, A.~Padalkar, A.~Lee, A.~Pooley, A.~Gupta, A.~Mandlekar, A.~Jain, \emph{et~al.}, ``Open x-embodiment: Robotic learning datasets and rt-x models: Open x-embodiment collaboration 0,'' in \emph{2024 IEEE International Conference on Robotics and Automation (ICRA)}.\hskip 1em plus 0.5em minus 0.4em\relax IEEE, 2024, pp. 6892--6903.

\bibitem{bu2025agibot}
Q.~Bu, J.~Cai, L.~Chen, X.~Cui, Y.~Ding, S.~Feng, S.~Gao, X.~He, X.~Hu, X.~Huang, \emph{et~al.}, ``Agibot world colosseo: A large-scale manipulation platform for scalable and intelligent embodied systems,'' \emph{arXiv preprint arXiv:2503.06669}, 2025.

\bibitem{wu2024robomind}
K.~Wu, C.~Hou, J.~Liu, Z.~Che, X.~Ju, Z.~Yang, M.~Li, Y.~Zhao, Z.~Xu, G.~Yang, \emph{et~al.}, ``Robomind: Benchmark on multi-embodiment intelligence normative data for robot manipulation,'' \emph{arXiv preprint arXiv:2412.13877}, 2024.

\bibitem{liu2023libero}
B.~Liu, Y.~Zhu, C.~Gao, Y.~Feng, Q.~Liu, Y.~Zhu, and P.~Stone, ``Libero: Benchmarking knowledge transfer for lifelong robot learning,'' \emph{Advances in Neural Information Processing Systems}, vol.~36, pp. 44\,776--44\,791, 2023.

\bibitem{mu2021maniskill}
T.~Mu, Z.~Ling, F.~Xiang, D.~Yang, X.~Li, S.~Tao, Z.~Huang, Z.~Jia, and H.~Su, ``Maniskill: Generalizable manipulation skill benchmark with large-scale demonstrations,'' in \emph{35th Conference on Neural Information Processing Systems Datasets and Benchmarks Track (Round 2)}, 2021.

\bibitem{gu2023maniskill2}
J.~Gu, F.~Xiang, X.~Li, Z.~Ling, X.~Liu, T.~Mu, Y.~Tang, S.~Tao, X.~Wei, Y.~Yao, \emph{et~al.}, ``Maniskill2: A unified benchmark for generalizable manipulation skills,'' \emph{arXiv preprint arXiv:2302.04659}, 2023.

\bibitem{james2020rlbench}
S.~James, Z.~Ma, D.~R. Arrojo, and A.~J. Davison, ``Rlbench: The robot learning benchmark \& learning environment,'' \emph{IEEE Robotics and Automation Letters}, vol.~5, no.~2, pp. 3019--3026, 2020.

\bibitem{zhang2025agentworld}
Y.~Zhang, Z.~Yu, J.~Lai, C.~Lu, and L.~Han, ``Agentworld: An interactive simulation platform for scene construction and mobile robotic manipulation,'' \emph{arXiv preprint arXiv:2508.07770}, 2025.

\bibitem{unitreerobotics_xr_teleoperate}
U.~Robotics, ``Xr teleoperation.'' \url{\\https://github.com/unitreerobotics/xr\_teleoperate}, 2024.

\bibitem{hunyuan3d2025hunyuan3d}
T.~Hunyuan3D, S.~Yang, M.~Yang, Y.~Feng, X.~Huang, S.~Zhang, Z.~He, D.~Luo, H.~Liu, Y.~Zhao, \emph{et~al.}, ``Hunyuan3d 2.1: From images to high-fidelity 3d assets with production-ready pbr material,'' \emph{arXiv preprint arXiv:2506.15442}, 2025.

\bibitem{moenne20243d}
N.~Moenne-Loccoz, A.~Mirzaei, O.~Perel, R.~de~Lutio, J.~Martinez~Esturo, G.~State, S.~Fidler, N.~Sharp, and Z.~Gojcic, ``3d gaussian ray tracing: Fast tracing of particle scenes,'' \emph{ACM Transactions on Graphics (TOG)}, vol.~43, no.~6, pp. 1--19, 2024.

\bibitem{brohan2022rt}
A.~Brohan, N.~Brown, J.~Carbajal, Y.~Chebotar, J.~Dabis, C.~Finn, K.~Gopalakrishnan, K.~Hausman, A.~Herzog, J.~Hsu, \emph{et~al.}, ``Rt-1: Robotics transformer for real-world control at scale,'' \emph{arXiv preprint arXiv:2212.06817}, 2022.

\bibitem{zitkovich2023rt}
B.~Zitkovich, T.~Yu, S.~Xu, P.~Xu, T.~Xiao, F.~Xia, J.~Wu, P.~Wohlhart, S.~Welker, A.~Wahid, \emph{et~al.}, ``Rt-2: Vision-language-action models transfer web knowledge to robotic control,'' in \emph{Conference on Robot Learning}.\hskip 1em plus 0.5em minus 0.4em\relax PMLR, 2023, pp. 2165--2183.

\bibitem{kim2025openvla}
M.~J. Kim, K.~Pertsch, S.~Karamcheti, T.~Xiao, A.~Balakrishna, S.~Nair, R.~Rafailov, E.~P. Foster, P.~R. Sanketi, Q.~Vuong, \emph{et~al.}, ``Openvla: An open-source vision-language-action model,'' in \emph{Conference on Robot Learning}.\hskip 1em plus 0.5em minus 0.4em\relax PMLR, 2025, pp. 2679--2713.

\bibitem{shi2025hi}
L.~X. Shi, B.~Ichter, M.~Equi, L.~Ke, K.~Pertsch, Q.~Vuong, J.~Tanner, A.~Walling, H.~Wang, N.~Fusai, \emph{et~al.}, ``Hi robot: Open-ended instruction following with hierarchical vision-language-action models,'' \emph{arXiv preprint arXiv:2502.19417}, 2025.

\bibitem{intelligence2504pi0}
P.~Intelligence, K.~Black, N.~Brown, J.~Darpinian, K.~Dhabalia, D.~Driess, A.~Esmail, M.~Equi, C.~Finn, N.~Fusai, \emph{et~al.}, ``$\pi$0. 5: a vision-language-action model with open-world generalization, 2025,'' \emph{URL https://arxiv. org/abs/2504.16054}, 2025.

\bibitem{black2024pi_0}
K.~Black, N.~Brown, D.~Driess, A.~Esmail, M.~Equi, C.~Finn, N.~Fusai, L.~Groom, K.~Hausman, B.~Ichter, \emph{et~al.}, ``$pi\_0 $: A vision-language-action flow model for general robot control,'' \emph{arXiv preprint arXiv:2410.24164}, 2024.

\bibitem{bjorck2025gr00t}
J.~Bjorck, F.~Casta{\~n}eda, N.~Cherniadev, X.~Da, R.~Ding, L.~Fan, Y.~Fang, D.~Fox, F.~Hu, S.~Huang, \emph{et~al.}, ``Gr00t n1: An open foundation model for generalist humanoid robots,'' \emph{arXiv preprint arXiv:2503.14734}, 2025.

\bibitem{zhao2020sim}
W.~Zhao, J.~P. Queralta, and T.~Westerlund, ``Sim-to-real transfer in deep reinforcement learning for robotics: a survey,'' in \emph{2020 IEEE symposium series on computational intelligence (SSCI)}.\hskip 1em plus 0.5em minus 0.4em\relax IEEE, 2020, pp. 737--744.

\bibitem{hofer2021sim2real}
S.~H{\"o}fer, K.~Bekris, A.~Handa, J.~C. Gamboa, M.~Mozifian, F.~Golemo, C.~Atkeson, D.~Fox, K.~Goldberg, J.~Leonard, \emph{et~al.}, ``Sim2real in robotics and automation: Applications and challenges,'' \emph{IEEE transactions on automation science and engineering}, vol.~18, no.~2, pp. 398--400, 2021.

\bibitem{todorov2012mujoco}
E.~Todorov, T.~Erez, and Y.~Tassa, ``Mujoco: A physics engine for model-based control,'' in \emph{2012 IEEE/RSJ international conference on intelligent robots and systems}.\hskip 1em plus 0.5em minus 0.4em\relax IEEE, 2012, pp. 5026--5033.

\bibitem{abeyruwan2023sim2real}
S.~W. Abeyruwan, L.~Graesser, D.~B. D’Ambrosio, A.~Singh, A.~Shankar, A.~Bewley, D.~Jain, K.~M. Choromanski, and P.~R. Sanketi, ``i-sim2real: Reinforcement learning of robotic policies in tight human-robot interaction loops,'' in \emph{Conference on Robot Learning}.\hskip 1em plus 0.5em minus 0.4em\relax PMLR, 2023, pp. 212--224.

\bibitem{radosavovic2023learning}
I.~Radosavovic, T.~Xiao, B.~Zhang, T.~Darrell, J.~Malik, and K.~Sreenath, ``Learning humanoid locomotion with transformers,'' \emph{CoRR}, 2023.

\bibitem{agarwal2023legged}
A.~Agarwal, A.~Kumar, J.~Malik, and D.~Pathak, ``Legged locomotion in challenging terrains using egocentric vision,'' in \emph{Conference on robot learning}.\hskip 1em plus 0.5em minus 0.4em\relax PMLR, 2023, pp. 403--415.

\bibitem{hu2023simulation}
X.~Hu, S.~Li, T.~Huang, B.~Tang, R.~Huai, and L.~Chen, ``How simulation helps autonomous driving: A survey of sim2real, digital twins, and parallel intelligence,'' \emph{IEEE Transactions on Intelligent Vehicles}, vol.~9, no.~1, pp. 593--612, 2023.

\bibitem{yuan2024robot}
Y.~Yuan, H.~Che, Y.~Qin, B.~Huang, Z.-H. Yin, K.-W. Lee, Y.~Wu, S.-C. Lim, and X.~Wang, ``Robot synesthesia: In-hand manipulation with visuotactile sensing,'' in \emph{2024 IEEE International Conference on Robotics and Automation (ICRA)}.\hskip 1em plus 0.5em minus 0.4em\relax IEEE, 2024, pp. 6558--6565.

\bibitem{lum2024dextrah}
T.~G.~W. Lum, M.~Matak, V.~Makoviychuk, A.~Handa, A.~Allshire, T.~Hermans, N.~D. Ratliff, and K.~Van~Wyk, ``Dextrah-g: Pixels-to-action dexterous arm-hand grasping with geometric fabrics,'' \emph{arXiv preprint arXiv:2407.02274}, 2024.

\bibitem{qureshi2024splatsim}
M.~N. Qureshi, S.~Garg, F.~Yandun, D.~Held, G.~Kantor, and A.~Silwal, ``Splatsim: Zero-shot sim2real transfer of rgb manipulation policies using gaussian splatting,'' \emph{arXiv preprint arXiv:2409.10161}, 2024.

\bibitem{liu2025fetchbot}
W.~Liu, Y.~Wan, J.~Wang, Y.~Kuang, X.~Shi, H.~Li, D.~Zhao, Z.~Zhang, and H.~Wang, ``Fetchbot: Object fetching in cluttered shelves via zero-shot sim2real,'' \emph{arXiv preprint arXiv:2502.17894}, 2025.

\bibitem{ye2022rcare}
R.~Ye, W.~Xu, H.~Fu, R.~K. Jenamani, V.~Nguyen, C.~Lu, K.~Dimitropoulou, and T.~Bhattacharjee, ``Rcare world: A human-centric simulation world for caregiving robots,'' in \emph{2022 IEEE/RSJ International Conference on Intelligent Robots and Systems (IROS)}.\hskip 1em plus 0.5em minus 0.4em\relax IEEE, 2022, pp. 33--40.

\bibitem{li2023behavior}
C.~Li, R.~Zhang, J.~Wong, C.~Gokmen, S.~Srivastava, R.~Mart{\'\i}n-Mart{\'\i}n, C.~Wang, G.~Levine, M.~Lingelbach, J.~Sun, \emph{et~al.}, ``Behavior-1k: A benchmark for embodied ai with 1,000 everyday activities and realistic simulation,'' in \emph{Conference on Robot Learning}.\hskip 1em plus 0.5em minus 0.4em\relax PMLR, 2023, pp. 80--93.

\bibitem{isaacsim4.5}
NVIDIA, ``Isaac sim 4.5 - robotics simulation and synthetic data generation,'' \emph{https: //developer.nvidia.com/isaacsim}, 2025.

\bibitem{rlbench}
S.~James, Z.~Ma, D.~R. Arrojo, and A.~J. Davison, ``Rlbench: The robot learning benchmark \& learning environment,'' \emph{IEEE Robotics and Automation Letters}, vol.~5, no.~2, pp. 3019--3026, 2020.

\bibitem{bigym}
N.~Chernyadev, N.~Backshall, X.~Ma, Y.~Lu, Y.~Seo, and S.~James, ``Bigym: A demo-driven mobile bi-manual manipulation benchmark,'' in \emph{8th Annual Conference on Robot Learning}.

\bibitem{behavior1k}
C.~Li, R.~Zhang, J.~Wong, C.~Gokmen, S.~Srivastava, R.~Mart{\'\i}n-Mart{\'\i}n, C.~Wang, G.~Levine, M.~Lingelbach, J.~Sun, \emph{et~al.}, ``Behavior-1k: A benchmark for embodied ai with 1,000 everyday activities and realistic simulation,'' in \emph{Conference on Robot Learning}.\hskip 1em plus 0.5em minus 0.4em\relax PMLR, 2023, pp. 80--93.

\bibitem{mimicgen}
A.~Mandlekar, S.~Nasiriany, B.~Wen, I.~Akinola, Y.~Narang, L.~Fan, Y.~Zhu, and D.~Fox, ``Mimicgen: A data generation system for scalable robot learning using human demonstrations,'' \emph{arXiv preprint arXiv:2310.17596}, 2023.

\bibitem{grutopia}
H.~Wang, J.~Chen, W.~Huang, Q.~Ben, T.~Wang, B.~Mi, T.~Huang, S.~Zhao, Y.~Chen, S.~Yang, \emph{et~al.}, ``Grutopia: Dream general robots in a city at scale,'' \emph{arXiv preprint arXiv:2407.10943}, 2024.

\bibitem{wu20253dgut}
Q.~Wu, J.~Martinez~Esturo, A.~Mirzaei, N.~Moenne-Loccoz, and Z.~Gojcic, ``3dgut: Enabling distorted cameras and secondary rays in gaussian splatting,'' \emph{Conference on Computer Vision and Pattern Recognition (CVPR)}, 2025.

\bibitem{kerbl20233d}
B.~Kerbl, G.~Kopanas, T.~Leimk{\"u}hler, and G.~Drettakis, ``3d gaussian splatting for real-time radiance field rendering.'' \emph{ACM Trans. Graph.}, vol.~42, no.~4, pp. 139--1, 2023.

\bibitem{jin2025artviparticulateddigitalassets}
\BIBentryALTinterwordspacing
Z.~Jin, Z.~Che, Z.~Zhao, K.~Wu, Y.~Zhang, Y.~Zhao, Z.~Liu, Q.~Zhang, X.~Ju, J.~Tian, Y.~Xue, and J.~Tang, ``Artvip: Articulated digital assets of visual realism, modular interaction, and physical fidelity for robot learning,'' 2025. [Online]. Available: \url{https://arxiv.org/abs/2506.04941}
\BIBentrySTDinterwordspacing

\bibitem{ycb}
B.~Calli, A.~Singh, A.~Walsman, S.~Srinivasa, P.~Abbeel, and A.~M. Dollar, ``The ycb object and model set: Towards common benchmarks for manipulation research,'' in \emph{2015 International Conference on Advanced Robotics (ICAR)}, 2015, pp. 510--517.

\bibitem{cadene2024lerobot}
R.~Cadene, S.~Alibert, A.~Soare, Q.~Gallouedec, A.~Zouitine, S.~Palma, P.~Kooijmans, M.~Aractingi, M.~Shukor, D.~Aubakirova, M.~Russi, F.~Capuano, C.~Pascal, J.~Choghari, J.~Moss, and T.~Wolf, ``Lerobot: State-of-the-art machine learning for real-world robotics in pytorch,'' \url{https://github.com/huggingface/lerobot}, 2024.

\end{thebibliography}


\end{document}